\documentclass[11pt]{article}

\usepackage[final]{acl}

\usepackage{amsmath}
\usepackage{booktabs}
\usepackage[table]{xcolor}
\usepackage{multirow}
\usepackage{tcolorbox}
\usepackage{enumitem}
\usepackage{subcaption}
\usepackage{xcolor}
\usepackage[nameinlink,noabbrev]{cleveref}
\usepackage{times}
\usepackage{latexsym}
\usepackage{float}

\usepackage[T1]{fontenc}

\usepackage[utf8]{inputenc}

\usepackage{microtype}

\usepackage{inconsolata}

\usepackage{graphicx}

\title{Refining and Reusing Annotation Guidelines for LLM Annotation}

\author{First Author \\
  Affiliation / Address line 1 \\
  Affiliation / Address line 2 \\
  Affiliation / Address line 3 \\
  \texttt{email@domain} \\\And
  Second Author \\
  Affiliation / Address line 1 \\
  Affiliation / Address line 2 \\
  Affiliation / Address line 3 \\
  \texttt{email@domain} \\}

\author{
    Kon Woo Kim\textsuperscript{1,2}\quad
    Jin-Dong Kim\textsuperscript{3}\quad
    Akiko Aizawa\textsuperscript{1,2} \\
    \textsuperscript{1} The Graduate University for Advanced Studies, SOKENDAI \\
    \textsuperscript{2} National Institute of Informatics (NII)\\
    \textsuperscript{3} BioData Science Initiative (BSI), National Institute of Genetics (NIG)\\
    \texttt{\{ken1204,aizawa\}@nii.ac.jp}\\
    \texttt{jdkim@dbcls.rois.ac.jp}\\
}

\begin{document}
\maketitle
\begin{abstract}
While Large Language Models (LLMs) demonstrate remarkable performance on zero-shot annotation tasks, they often struggle with the specialized conventions of gold-standard benchmarks. We propose the systematic reuse and refinement of annotation guidelines as an alignment mechanism, introducing an iterative moderation framework that simulates the early phases of annotation projects. We evaluate three hypotheses: (1) the efficacy of guideline integration, (2) the advantage of reasoning-optimized models, and (3) the viability of moderation under minimal supervision. Testing across biomedical NER tasks (NCBI Disease, BC5CDR, BioRED) with three LLM families (GPT, Gemini, DeepSeek), our results empirically confirm all three hypotheses. While the iterative moderation framework shows good potential in effectively refining guidelines, our analysis also reveals substantial room for improvement.
\end{abstract}

\section{Introduction}
\label{sec:introduction}
Text annotation serves as the foundation for various text processing tasks, including semantic search and text mining.
Recently, zero-shot or few-shot prompting of Large Language Models (LLMs) has demonstrated impressive performance in general labeling tasks. However, when annotations must serve specific downstream requirements, regulating LLM outputs to adhere to these concrete constraints becomes non-trivial.

Traditionally, in the corpus annotation community, annotation guidelines have been the primary mechanism for regulating the behavior of human annotators. Even when domain experts possess sufficient background knowledge, the realization of that knowledge into specific annotations can vary significantly when they encounter actual textual expressions. Discrepancies often arise regarding the scope of target entities, the resolution of ``gray-zone'' cases, and the precise determination of span boundaries. Annotation guidelines exist to standardize these behaviors, preventing what would otherwise be a stochastic or inconsistent process. Consequently, these guidelines encode not only high-level conceptual goals, but also the granular, task-specific requirements necessary to optimize annotations for practical applications.
This is precisely where the value of annotation guidelines lies: they are the result of rigorous optimization efforts designed to translate downstream application requirements into concrete labeling rules. As such, they constitute an invaluable resource for any text annotation effort.

For LLM-based annotation, prompting has become the primary mechanism for generating annotations; however, the challenge lies in effectively encoding these specialized requirements within the prompt. This work addresses this challenge by testing the following hypothesis:
\begin{quote}
H1: The incorporation of annotation guidelines into the prompt will improve the LLM-based annotations.
\end{quote}

Guideline-driven annotation requires sophisticated cognitive processing, as the application of complex rules to diverse textual contexts can be viewed as a deductive reasoning process. Therefore, it is expected that models with superior reasoning capabilities will perform better on this task. This leads to our second hypothesis:
\begin{quote}
H2: Reasoning models will outperform their non-reasoning counterparts in guideline-driven annotation tasks.
\end{quote}

Guidelines are often incomplete. This is evident in traditional corpus projects where new annotators require a ``training'' phase. During this phase, annotators annotate a small set of documents to identify discrepancies against gold annotations. This process, known as moderation, results in either the refinement of the annotator’s understanding or the clarification of the guidelines themselves. Notably, this process is typically conducted with minimal supervision. Because guidelines are inherently iterative and subject to initial ambiguity, our final hypothesis is:
\begin{quote}
H3: Annotation guidelines can be effectively improved through a moderation process with minimal supervision.
\end{quote}
Note that the task of guideline refinement itself can be framed as a process of inductive reasoning.
This dual requirement for both deductive and inductive reasoning further motivates our second hypothesis.

The contributions of this work are threefold:
\begin{enumerate}
    \item We provide an empirical validation of the three hypotheses concerning the role of guidelines, reasoning capabilities, and moderation in LLM-based annotation.
    \item We introduce an iterative moderation framework for guideline refinement and demonstrate its efficacy in aligning LLM outputs with gold-standard conventions under minimal supervision.
    \item We provide a detailed qualitative analysis of the refinement process, offering insights into the evolution of guidelines and the specific linguistic discrepancies addressed during iteration to facilitate future research.
\end{enumerate}

\section{Related Work}

\paragraph{Biomedical corpora and document-level annotation.}
The NCBI Disease Corpus \citep{dogan-etal-2014-ncbi} provides manually curated disease mentions and concept normalization on PubMed abstracts,
while BC5CDR \citep{li-etal-2016-bc5cdr} supports chemical and disease entities with relation annotations for the BioCreative CDR task.
BioRED \citep{luo2022biored} extends document-level relation extraction with multiple entity types and relation pairs.
These datasets exemplify annotation projects in which detailed human-written guidelines and moderation are essential for handling boundary cases and ambiguity
\citep{islamaj2020teamtat}.
We focus on \emph{NER-only} (exact span+type), as guideline refinement in our setting primarily targets mention boundaries and type assignments.

\paragraph{LLMs as annotators and guideline followers.}
Until recently, many annotation projects have been implemented through crowd-sourcing due to faster and cheaper execution \citep{callison2009fast}.
Beyond traditional annotation pipelines, prior work has explored LLMs as substitutes or complements to human annotators \citep{wang2021want,ding2023gpt}.
Some studies show that ChatGPT can match or exceed crowd-worker quality on several text-annotation tasks, and motivates LLM-based labeling pipelines \citep{gilardi-etal-2023-chatgpt,zhu2023can}.
However, reliable deployment depends on whether LLMs can adhere to detailed labeling rules rather than generic task descriptions. Another study investigates guideline-following behavior across domains, finding that performance can be brittle and sensitive to guideline structure, ambiguity, and edge cases \citep{fonseca-cohen-2024-large}.

\paragraph{Evaluation with LLMs as judges and evaluator.}
Several studies show the potential of LLMs as judges for open-ended outputs \citep{zhu2023judgelm} and discuss key biases, such as position and verbosity effects \citep{zheng-etal-2023-mtbench}.
Subsequent work explores LLMs as structured evaluator rather than just scorers \citep{liu2023g}.
Our setting differs in that the judge is constrained by explicit domain guidelines and disagreement evidence, and is tasked with producing actionable guideline refinements rather than a scalar preference score.

\paragraph{Utilization of Guidelines for LLMs.}
Some studies reveal that using guidelines helps LLMs with consistency by providing clear definitions \citep{huang2025guidener} and instructions \citep{srivastava2025instruction}. 
A case study illustrates that LLMs help identify guideline issues and propose refinements and suggests that disagreement patterns can be exploited to iteratively strengthen guidelines \citep{bibal-etal-2025-automating}.
Another recent work investigates whether existing human annotation guidelines can be directly repurposed for LLM-based annotation, highlighting both the promise of guideline-driven prompting and the difficulty of faithfully translating complex domain-specific rules into model behavior \citep{kim2025repurposing}.
Recent work has also explored the development of LLM-assisted annotation-scheme \citep{fernandes2025human}. Whereas that line of research focuses on constructing annotation schemes, our work focuses on the iterative refinement and reuse of existing human-written guidelines under minimal supervision.

\begin{figure*}[t]
    \centering
    \includegraphics[
        width=0.9\linewidth,
    ]{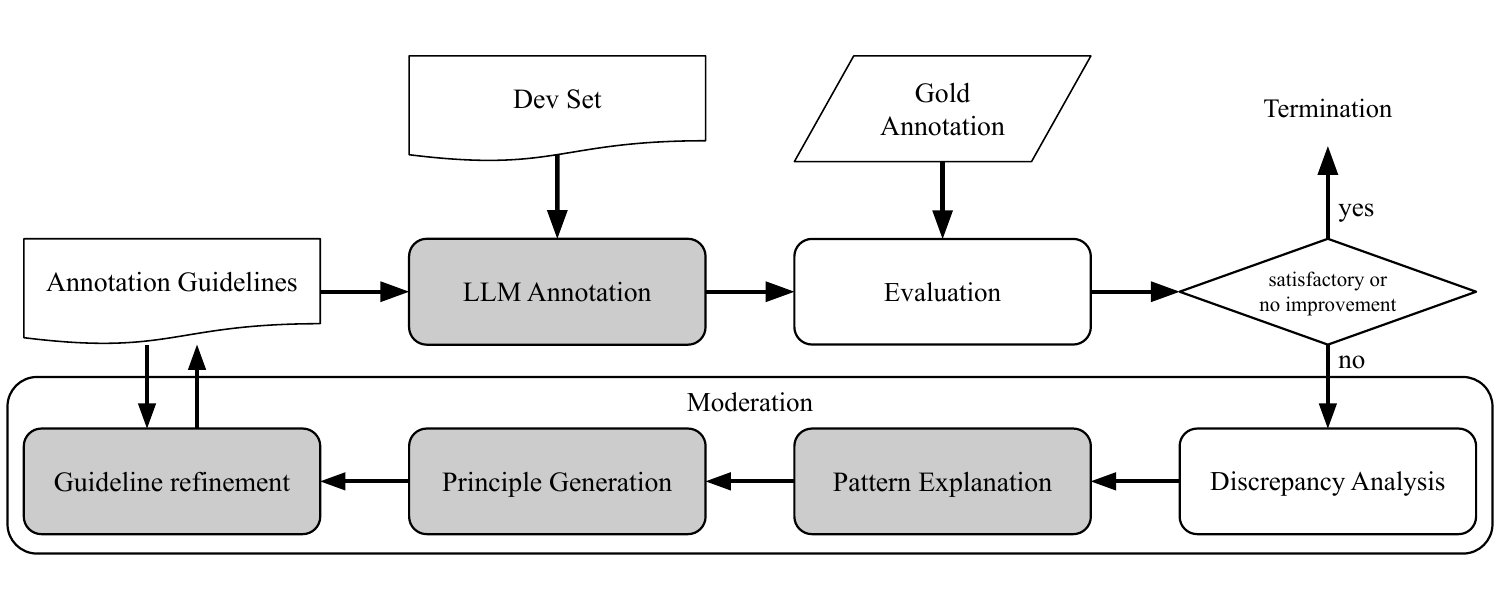}
    \caption{Overview of the iterative moderation framework. Given the current annotation guidelines, an LLM annotates the development set, and the resulting annotations are evaluated against gold annotations using strict span-and-type matching. If performance is satisfactory or no longer improves, the process terminates. Otherwise, discrepancy analysis identifies the dominant error pattern, which is passed through three moderation stages: pattern explanation, principle generation, and guideline refinement. The refined guideline is then reused in the next iteration.}
    \label{fig:overview}
\end{figure*}

\section{Methodology}

\begin{figure*}[!t]
\centering
\fbox{%
\begin{minipage}{0.97\textwidth}
\small

\textbf{(A) Pattern Explanation (Dominant Error Pattern Group)} \\
\vspace{0.25em}
\begin{quote}
\textit{Pattern Name:} Phenotypic feature-list contexts suppress \texttt{DiseaseClass} tagging. \\
\textit{Confusion Trigger:} Condition terms used as items in dependent ``feature lists'' (e.g., \texttt{with/of} complements,
or coordinated arguments of report/causal verbs) are treated as mere symptoms/processes and dropped. \\
\textit{Contrastive Evidence:} True positives concentrate in definitional disease-head frames (e.g., ``X is an autosomal recessive disorder''),
where an explicit disease-head noun cues \texttt{DiseaseClass}. \\
\textit{Rule (Proposed):} If an NP denotes a clinical condition and appears as an item in a dependent feature-list
(e.g., within a \texttt{with/of} complement or a coordinated report/causal frame), annotate each conjunct as \texttt{DiseaseClass};
do not suppress solely due to list presentation.
\end{quote}

\vspace{0.45em}
\hrule
\vspace{0.45em}

\textbf{(B) Principle Generation (Single General Principle)} \\
\vspace{0.25em}
\begin{quote}
\textbf{IF} a noun phrase denoting a clinical condition appears as an item in a dependent ``feature-list'' construction
(e.g., as a conjunct within a \texttt{with/of} complement of a patient/syndrome noun, or as a coordinated argument governed by
observation/reporting/causation verbs), \\
\textbf{THEN} annotate that phrase (and propagate to all coordinated conjuncts) as \texttt{DiseaseClass} using the minimum necessary span; \\
\textbf{EXCEPT} when the condition phrase functions only as a modifier of a non-disease head (e.g., gene/locus/region/protein/variant)
or is a bare/general process term that should not be annotated as a disease.
\end{quote}

\vspace{0.45em}
\hrule
\vspace{0.45em}

\textbf{(C) Guideline Refinement (Inserted Rule; Truncated)} \\[-0.2em]
\vspace{0.25em}

\begin{minipage}{0.97\textwidth}
\footnotesize
\setlength{\parskip}{0.25em}

\textbf{\#4. Annotate coordinated headless phenotype/pathology complements as \texttt{DiseaseClass}}

When a coordinated sequence of \emph{headless} phenotype/pathology noun phrases
(e.g., \texttt{anemia}, \texttt{lymphadenopathy}, \texttt{rash})
functions as the complement of a clinical-description or diagnostic construction,
annotate \emph{each coordinated item} as \texttt{DiseaseClass} using the minimum span.

\textbf{Applicable constructions include:}
\begin{itemize}[leftmargin=1.1em, itemsep=0.15em, topsep=0.15em]
  \item Patient/disease-centered descriptions
  (e.g., \texttt{patients with ...}, \texttt{characterized by ...}, \texttt{presenting with ...}).
  \item Syndrome-level frames introducing a phenotypic list
  (e.g., \texttt{a syndrome characterized by ...}).
\end{itemize}

\textbf{Do not apply when:}
\begin{itemize}[leftmargin=1.1em, itemsep=0.15em, topsep=0.15em]
  \item Items are bare general or process terms
  (e.g., \texttt{symptoms}, \texttt{carcinogenesis}).
  \item The coordination is governed by an overt disease head
  (e.g., \texttt{breast and ovarian cancer}; follow \texttt{CompositeMention}).
\end{itemize}

\textbf{Note.}
This rule refines \texttt{DiseaseClass} only and does not affect \texttt{SpecificDisease}
or disease \texttt{Modifier} handling (e.g., \texttt{ALPS} vs.\ \texttt{ALPS phenotype}). \\
(\textit{Truncated due to space limit. Full text is provided in Figure~\ref{fig:appendix_guideline_rule4_latex}.})
\end{minipage}

\end{minipage}
}

\caption{
Example of LLM-simulated moderation for DiseaseClass guideline refinement.
The moderator identifies a recurring failure where DiseaseClass mentions embedded in dependent feature-list contexts
(e.g., \texttt{with/of} complements and coordinated report/causal frames) are suppressed, generates an actionable principle,
and inserts a targeted guideline rule in order to reduce DiseaseClass false negatives. The following guideline from (C) is inserted between Section 3 (Modifiers) and Section 4 (Duplicate mentions) in the ``What to Annotate'' section of the original human annotation guidelines, as shown in Figure~\ref{fig:human_guidelines_example}.\\
}
\label{fig:moderation_example_disease}
\end{figure*}

\subsection{Task formulation}
The objective of the task is to adapt existing annotation guidelines - originally developed for human annotators - to regulate the performance of LLMs.

In traditional annotation projects, guidelines are established during a pilot phase where multiple human annotators label the same document set. During this \emph{moderation} stage, \emph{Inter-Annotator Agreement (IAA) rate} is measured, and guidelines are iteratively refined to resolve ambiguities and improve consistency. If new annotators join mid-project, they undergo a \emph{training} phase: they annotate a small set of documents with gold annotations, and their performance is evaluated against the gold annotations.
If the IAA rate is not high enough, the annotators are either retrained or, preferably, the guidelines are refined to ensure better reproducibility and objective clarity.

In this work, we explore the potential of using existing annotation guidelines to enable LLMs to reproduce annotations based on the guidelines.
We frame this process as being equivalent to training new human annotators through iterative guideline refinement, however with LLM annotators instead of human annotators.
As outlined in Figure~\ref{fig:overview}, the proposed framework operates as an iterative cycle encompassing (1) annotation, (2) evaluation, and (3) moderation.

\subsection{LLM-based Annotation}
At the iteration $k$, a LLM annotator is prompted to annotate the documents in the development set $D$ using the current set of annotation guidelines $G_k$. This process yields a set of predicted annotations $A_k$.

\subsection{Evaluation}
To assess the LLM's performance, the predicted annotations $A_k$ are compared against the gold annotations $A_g$.
As a measure of IAA rate, the F1-score is calculated based on a \emph{strict matching criterion}, which requires exact match of both the entity boundary and the entity type.

After evaluation, the termination criteria are assessed. The iteration concludes if either:
\begin{enumerate}
\item The alignment reaches a predefined quality threshold ($\text{IAA}_k \geq \tau$).
\item For refined versions ($k > 0$), the most recent refinement fails to yield a performance gain ($\text{IAA}_k \leq \text{IAA}_{k-1}$)
\end{enumerate}

\subsection{Moderation}
\label{sec:moderation}
When the evaluation results fall below the required threshold,
a moderation process is triggered to refine the guidelines.
The LLM annotator then re-annotates the development set using the updated guidelines,
forming an iterative loop.
This process continues until the desired performance is achieved or a termination condition is met.
Specifically, the loop terminates if a moderation cycle fails to yield measurable improvement,
in which case the most recent refinement is discarded to prevent performance degradation.

At each iteration $k$, the moderation takes two inputs:
(1) the current set of guidelines $G_k$, and
(2) a prioritized group of discrepancy patterns identified during the analysis phase.
A refined version of the guidelines $G_{k+1}$ is then generated,
through a process we call \emph{moderation}.
The moderation process is designed to take three steps as shown in Figure~\ref{fig:moderation_example_disease}:
(1) pattern explanation, (2) principle generation, and (3) guideline refinement.

\subsubsection{Discrepancy Analysis}
\label{sec:analysis}

To facilitate guideline refinement, we collect and analyze all discrepancies between $A_k$ and $A_g$.
While the primary evaluation is strict, we apply a \emph{soft matching criterion} (requiring at least one character of overlap) to categorize errors for finer-grained diagnostic analysis.
Each discrepancy case is assigned to one of four mutually exclusive categories, prioritized in the following order:
\begin{enumerate}
  \item \textbf{Label mismatch}: A gold entity and a predicted entity overlap in text (at least one character) but have different types.
  \item \textbf{Boundary mismatch}: a gold entity and a predicted entity overlap in text and have the same type, but their start/end offsets differ.
  \item \textbf{False Negative (FN)}: a gold entity with zero character overlap with any predicted entity.
  \item \textbf{False Positive (FP)}: a predicted entity with zero character overlap with any gold entity.
\end{enumerate}
Discrepancy cases are grouped based on their predicted and gold label pairs,
and the most frequent group is selected as the target of moderation.

Each discrepant case is represented by
(1) the mention string(s),
(2) the surrounding context (window size = 60),
(3) the types of predicted and gold entities, and
(4) the discrepancy category.

Here, the entity labels (e.g., \textsc{DiseaseClass}, \textsc{SpecificDisease}, \textsc{Modifier}) are defined by the dataset schema, whereas the discrepancy categories above are analytic categories over prediction--gold comparisons. These two levels should be distinguished: the former specifies what can be annotated, while the latter specifies how annotation disagreements are classified. In addition, label mismatches are treated as directional by
grouping cases according to gold--predicted label pairs, rather than collapsing them into symmetric confusion types.

\subsubsection{Pattern Explanation}
The goal of this step is to generate an explanation for the targeted discrepancy pattern group. To do so, we provide (a) discrepancy examples from the selected group and (b) verified true positives that are correctly handled under $G_k$. The model is then asked to compare these two sets contrastively in order to identify annotation-discriminative evidence that separates the discrepant cases from the verified true positives. 

This evidence may concern (i) semantic cues that affect label assignment or (ii) structural or positional cues that affect span decisions, such as coordination boundaries, attachment, nesting, or contiguity. The purpose of this step is not to introduce a new linguistic taxonomy but rather to summarize the specific evidence that appears to govern the annotation distinction in the current discrepancy group. The model outputs exactly one pattern insight in a fixed format so that the result can be used consistently in the subsequent moderation stages.

\subsubsection{Principle Generation}
This step generates a generalized moderation principle from the pattern explanation produced in the previous step.
In this step, the extracted pattern insight is converted into one actionable rule suitable for human annotators and future LLM annotators.
We cast the model as an \textit{AI moderator} and provide entity definitions, the linguistic analysis, and the current guideline.
The moderator must synthesize a single general principle in a strict IF/THEN form and include an explicit negative constraint to specify when the rule \emph{does not} apply to avoid over-correction and false positives.
To improve generalization, we require abstract phrasing rather than instantiations tied to specific tokens.
Finally, the model checks consistency against existing entity definitions and guideline rules.

\subsubsection{Guideline Refinement}
Finally, this step integrates the principle generated in the previous step into the current set of guidelines $G_k$,
and generates a revised version of guidelines $G_{k+1}$.
The primary role of this step is to maintain the integrity during the revision.
We provide the current guideline, the newly generated principle, and the discrepancy context.
We include verified examples (true positives) as in-prompt checks: the model is instructed not to introduce changes that would flip these known-correct cases.
To preserve document integrity, the model outputs the full updated guideline text without omission or summarization, while maintaining the original formatting and structure.

\section{Experimental Setup}

\subsection{Implementation}
We explored biomedical named entity recognition (NER) tasks as a testbed of the methodology.
For NER tasks, an individual annotation is represented by triplets of the character offsets of the beginning and end of the entity, and the type of the entity.

We used the PubAnnotation framework for management, evaluation, visualization of annotations \citep{kim2019open}.
Accordingly, the PubAnnotation JSON format is used to represent annotations.
All annotations and evaluation results are stored in PubAnnotation for reproducibility and are also available at \url{https://github.com/KonWooKim/llm-guideline-moderation}.

\subsection{Datasets}
We evaluate on biomedical datasets with established human annotation guidelines:
\begin{itemize}
  \item \textbf{NCBI Disease}: PubMed abstracts annotated for disease mentions.
  \item \textbf{BC5CDR}: PubMed abstracts annotated for Chemical and Disease mentions (NER-only).
  \item \textbf{BioRED}: PubMed abstracts annotated with biomedical entity types (and relations); we focus on NER-only entity denotations.
\end{itemize}
To provide a rough sense of benchmark difficulty, prior work reports official-test performance of 92.28 F1 on NCBI Disease, 93.74 on BC5CDR-Chemical, and 90.69 on BC5CDR-Disease \citep{jiang-etal-2021-named}, and around 89 F1 on BioRED \citep{luo2022biored}. These figures are not directly comparable to our setup, but provide context for interpreting the dataset.

For reference, per-type entity counts for the 100-document evaluation sets are reported in Appendix Table~\ref{tab:entity_counts}.

Since our primary objective is to evaluate the feasibility of inducing high-level annotation principles,
rather than achieving state-of-the-art performance,
we departed from the standard train/dev/test splits provided by the original dataset authors.
Instead, we adopted a ``low-resource'' approach to test the hypothesis that high-level guidelines can be induced from minimal supervision.

Specifically, we randomly sampled 10 documents from the original training split of each benchmark to serve as the \emph{development dataset} for iterative moderation. This choice was intended to simulate the pilot stage of real annotation workflows, where guideline development is often conducted on a small representative set before large-scale annotation begins. Our design therefore intentionally mirrors a practically constrained early-stage setting in which expert supervision is limited and guideline ambiguities must be resolved from sparse evidence.

To offset the instability of such a small refinement set, we separately prepared a larger evaluation set of 100 documents from the original development split as our \emph{evaluation dataset} for final performance assessment. In other words, guideline refinement is performed on 10 documents, whereas the performance is measured on a distinct 100-document evaluation set.

For NCBI Disease and BioRED, we used their respective development splits in full, as both contain exactly 100 documents.
For BC5CDR, which provides a larger development split of 500 documents, we randomly sampled 100.
Finally, we intentionally withheld the original test sets
to preserve their integrity for future benchmarking and to ensure that our refinement process remains strictly separated from the final unseen data.

\subsection{Selection of Large Language Models}
We selected three representative, state-of-the-art Large Language Model (LLM) families for our experiments:
\emph{GPT}, \emph{Gemini}, and \emph{DeepSeek}.
Our model selection was guided by two criteria: (1) sufficient reasoning capacity for stable adherence to detailed annotation guidelines, and (2) coverage of comparable SOTA LLM families that provide both reasoning-oriented and non-reasoning variants for controlled comparison.
Within each family, we distinguish between non-reasoning and reasoning models.

Preliminary pilot experiments with smaller local models showed unsatisfactory baseline performance, unstable adherence to refined guidelines, and no consistent gains after moderation. Thus, we excluded such models from the main experiments and focused on model families that allow a more reliable comparison of guideline
integration, reasoning, and moderation effects (See Table~\ref{tab:pilot_local_models}).

Our primary hypothesis is that the iterative process of guideline refinement and application requires high-order reasoning capabilities.
Specifically, synthesizing new guidelines from discrepancies requires inductive reasoning, whereas applying those guidelines to specific instances requires deductive reasoning.
To test this hypothesis, our first experiment compares the performance of non-reasoning and reasoning models within the same families to evaluate their relative effectiveness in this iterative loop.
The specific models and configurations are as follows:
\begin{itemize}
  \item \textbf{GPT}: \texttt{gpt-5-2025-08-07} with \texttt{reasoning\_effort} $\in \{\texttt{low}, \texttt{high}\}$.
\item \textbf{Gemini}: \texttt{gemini-2.5-pro} with \texttt{thinking\_budget} $\in \{\texttt{min}, \texttt{max}\}$.
  \item \textbf{DeepSeek}: \texttt{deepseek-chat} (non-reasoning) vs.\ \texttt{deepseek-reasoner} (reasoning).
\end{itemize}
We ran all models with default hyperparameter settings, as reasoning models expose non-comparable configuration options (e.g., GPT-5 does not support \texttt{temperature}, whereas Gemini does).

\subsection{Design of Experiments}
We designed our experiments to explore the three hypotheses presented in Section~\ref{sec:introduction}.
Accordingly, we compare three approaches:
\begin{enumerate}
  \item \textbf{Prompt-only} (baseline): minimal task instruction, annotation entities, and required output schema (see Figure~\ref{fig:prompt_template}).
  \item \textbf{Original-guidelines}: original human guideline text with light formatting (see Figure~\ref{fig:human_guidelines_example}).
  \item \textbf{Guideline-refinement}: discrepancy-driven guideline refinement.
\end{enumerate}

For the second approach, we experimented with both reasoning and non-reasoning models of each LLM family,
and, as reported in Section \ref{sec:reasoning-vs-non-reasoning}, 
across all datasets, reasoning models consistently outperform their non-reasoning counterparts (Table~\ref{tab:reasoning_effect_strict}).
Based on this result, we use only the reasoning models in the third approach.

\section{Results}
\label{sec:results}

\begin{table*}[t]
\centering
\small
\setlength{\tabcolsep}{4pt}

\resizebox{\textwidth}{!}{%
\begin{tabular}{llc|cccc|cccc|cccc}
\toprule
Dataset (\#Entity) & Model & \#Iters
& \multicolumn{4}{c}{S}
& \multicolumn{4}{c}{G ($\Delta$)}
& \multicolumn{4}{c}{M ($\Delta$)} \\
\cmidrule(lr){4-7} \cmidrule(lr){8-11} \cmidrule(lr){12-15}
& & & P & R & F1 & TP & P & R & F1 & TP & P & R & F1 & TP \\
\midrule
\multirow{3}{*}{NCBI (791)}
 & GPT-5 & 3
 & 0.45 & 0.48 & 0.46 & 378
 & 0.78 & 0.68 & 0.73 (+0.27) & 540
 & 0.82 & 0.71 & \textbf{0.76} (\textbf{+0.03}) & 565 \\
 & Gemini & 5
 & 0.36 & 0.47 & 0.40 & 369
 & 0.69 & 0.57 & 0.63 (+0.23) & 453
 & 0.72 & 0.61 & \textbf{0.66} (\textbf{+0.03}) & 479 \\
 & DeepSeek & 2
 & 0.32 & 0.30 & 0.31 & 236
 & 0.72 & 0.45 & 0.55 (+0.24) & 356
 & 0.71 & 0.47 & \textbf{0.56} (\textbf{+0.01}) & 369 \\
\midrule
\multirow{3}{*}{BC5CDR (2,146)}
 & GPT & 1
 & 0.84 & 0.78 & 0.80 & 1,664
 & 0.89 & 0.81 & 0.85 (+0.05) & 1,735
 & 0.92 & 0.81 & \textbf{0.86} (\textbf{+0.01}) & 1,737 \\
 & Gemini & 1
 & 0.74 & 0.63 & 0.68 & 1,359
 & 0.84 & 0.68 & 0.76 (+0.08) & 1,469
 & 0.86 & 0.70 & \textbf{0.77} (\textbf{+0.01}) & 1,503 \\
 & DeepSeek & 1
 & 0.80 & 0.45 & 0.58 & 968
 & 0.89 & 0.50 & 0.64 (+0.06) & 1,072
 & 0.86 & 0.52 & \textbf{0.65} (\textbf{+0.01}) & 1,119 \\
\midrule
\multirow{3}{*}{BioRED (3,531)}
 & GPT-5 & 2
 & 0.75 & 0.74 & 0.74 & 2,598
 & 0.81 & 0.72 & 0.76 (+0.02) & 2,548
 & 0.82 & 0.81 & \textbf{0.82} (\textbf{+0.06}) & 2,871 \\
 & Gemini & 1
 & 0.62 & 0.60 & 0.61 & 2,111
 & 0.74 & 0.61 & 0.67 (+0.06) & 2,137
 & 0.71 & 0.69 & \textbf{0.69} (\textbf{+0.02}) & 2,371 \\
 & DeepSeek & 1
 & 0.71 & 0.33 & 0.45 & 1,179
 & 0.77 & 0.41 & 0.53 (+0.08) & 1,442
 & 0.76 & 0.42 & \textbf{0.54} (\textbf{+0.01}) & 1,480 \\
\bottomrule
\end{tabular}%
}

\caption{
Performance of LLM annotation in three approaches: simple prompting (S), guideline (G), and moderation (M).
Precision (P), Recall (R), F1-score (F1), and the number of true positives (TP) are reported.
\#Iters indicates the number of moderation iterations.
}
\label{tab:grand_guideline_moderation_prf_tp}
\end{table*}

\subsection{With vs. Without Annotation Guidelines}
\begin{table}[t]
\centering
\small
\setlength{\tabcolsep}{5pt}
\begin{tabular}{llccc}
\toprule
Dataset & Model & Non-Reason & Reason & $\Delta$ \\
\midrule
\multirow{3}{*}{NCBI} 
& GPT      & 0.69 & \textbf{0.73} & +0.04 \\
& Gemini   & 0.48 & \textbf{0.63} & +0.15 \\
& DeepSeek & 0.29 & \textbf{0.55} & +0.26 \\
\midrule
\multirow{3}{*}{BC5CDR} 
& GPT      & 0.78 & \textbf{0.85} & +0.07 \\
& Gemini   & 0.70 & \textbf{0.76} & +0.06 \\
& DeepSeek & 0.57 & \textbf{0.64} & +0.07 \\
\midrule
\multirow{3}{*}{BioRED} 
& GPT      & 0.72 & \textbf{0.76} & +0.04 \\
& Gemini   & 0.66 & \textbf{0.67} & +0.01 \\
& DeepSeek & 0.43 & \textbf{0.53} & +0.10 \\
\bottomrule
\end{tabular}
\caption{
Comparison of (Non)-reasoning models.
}
\label{tab:reasoning_effect_strict}
\end{table}

As reported in the \emph{S} and \emph{G} column groups of Table~\ref{tab:grand_guideline_moderation_prf_tp},
providing annotation guidelines substantially improved the performance across all datasets and models.
This observation confirms the hypothesis that the provision of guidelines enhances the performance of LLM annotators.

\subsection{Reasoning vs. Non-reasoning models}
\label{sec:reasoning-vs-non-reasoning}

As reported in Table~\ref{tab:reasoning_effect_strict}, reasoning models consistently achieved higher performance than non-reasoning models across all datasets, with the approach of guideline provision.
This observation confirms the hypothesis that reasoning capability is essential for guideline-driven annotation.
A similar but weaker trend is also observed in preliminary prompt-only comparisons (Appendix Table~\ref{tab:prompt_reasoning}).

\subsection{With vs. Without Guideline Refinement}
As reported in the \emph{M} column group of Table~\ref{tab:grand_guideline_moderation_prf_tp},
the guideline refinement via the moderation approach yielded further performance gains.
While these improvements are marginal (typically +0.01 -- 0.03 in F1), they are observed consistently across all datasets and models.
These results demonstrate the potential of the moderation-based guideline refinement approach to enhance LLM annotation performance. 
However, the modest nature of the gains also indicates substantial room for improving both the robustness of the moderation procedure and the scope of discrepancy patterns it can reliably capture.

Because these gains are small in absolute terms, we additionally assessed their reliability using paired document-level bootstrap resampling and approximate randomization tests with one-sided hypotheses. The results show a model-dependent pattern: Gemini exhibits statistically significant gains across datasets, GPT shows the strongest evidence on BioRED and borderline evidence on NCBI, and DeepSeek shows positive but non-significant trends. Overall, these findings suggest that the effect of moderation is modest but partly reliable and becomes more consistently observable in stronger reasoning models (See Appendix Table~\ref{tab:significance_all}).

\subsection{Analysis of Iterative Moderation}
Figure~\ref{fig:ncbi_gpt_iters_1x4} illustrates the evolution of discrepancy patterns across moderation iterations performed by GPT on the NCBI Disease development dataset.  This analysis focuses on the 10-document refinement set and is intended as a diagnostic illustration of early-stage moderation, rather than as a statistically representative estimate of overall performance.
In this setup, the iterative process concluded after four iterations.
At each step, the LLM annotator's predictions ($A_0, \dots, A_3$) are compared against the gold annotations ($A_g$) to summarize how the remaining discrepancy structure changes over time. For instance, it demonstrates how the refinement process progressively resolves targeted label discrepancies while also revealing the secondary effects of these updates on other annotation cases.

In the initial iteration, the prioritized discrepancy pattern involved missed \textsc{DiseaseClass} mentions
(Predicted: No Entity, Gold: \textsc{DiseaseClass}),
which occurred with a frequency of 7.
Figure~\ref{fig:moderation_example_disease} illustrates how the annotation guidelines are refined in response to this pattern, and the corresponding matrix in Figure~\ref{fig:ncbi_gpt_iters_1x4} shows that this discrepancy type was largely resolved from 7 to 1 after the first iteration.

Importantly, these mismatch patterns are directional rather than symmetric. In this case, the main effect of moderation is not a uniform reduction of all errors, but a targeted recovery of previously missed or misclassified \textsc{DiseaseClass} mentions. At the same time, refinement can introduce new false positives, indicating a precision--recall rebalancing rather than monotonic improvement. This behavior is consistent with the small but systematic gains reported in Section~5.3 and helps explain why moderation may improve alignment while still producing new errors.

These secondary effects are not exclusively negative. For instance, five label-mismatch cases in which entities were incorrectly labeled as \textsc{Modifier} instead of the gold label \textsc{SpecificDisease} were resolved after the first iteration, suggesting that a rule introduced for one discrepancy group can propagate to related cases. Figure~\ref{fig:qual_fn} shows some positive changes (true positives) caused by the guideline refinement,
while Figure~\ref{fig:qual_fp} highlights a newly introduced false case (false positive) as a trade-off.
Detailed before/after examples with gold and predicted annotations are available in Appendix Figure~\ref{fig:detailed_examples}.

\begin{figure*}[t]
\centering
\scriptsize
\setlength{\tabcolsep}{3.2pt}
\renewcommand{\arraystretch}{1.03}

\begin{subfigure}[t]{0.245\textwidth}
\centering
\caption{Iter 0 ($G_0$)}
\label{fig:ncbi_gpt_iter0_1x4}
{\scriptsize \textbf{Gold} $\downarrow$ \hspace{0.4em} \textbf{Pred} $\rightarrow$ \par}
\vspace{0.15em}
\begin{tabular}{@{}c@{\hspace{0.45em}}c@{}}
\begin{tabular}{c|ccccc}
\toprule
 & C & D & M & S & O \\
\midrule
C & 0 & 0 & 0 & 0 & 0 \\
D & \cellcolor{red!46}3 & 0 & \cellcolor{red!22}1 & 0 & \cellcolor{red!80}7 \\
M & \cellcolor{red!22}1 & 0 & 0 & 0 & \cellcolor{red!22}1 \\
S & 0 & \cellcolor{red!58}5 & \cellcolor{red!58}5 & 0 & \cellcolor{red!46}3 \\
O & 0 & \cellcolor{red!22}1 & 0 & \cellcolor{red!22}1 & 0 \\
\bottomrule
\end{tabular}
&
{\scriptsize
\fbox{\begin{tabular}{@{}l@{}}
\textbf{TP}\\
C: 0\\
D: 12\\
M: 14\\
S: 34
\end{tabular}}}
\end{tabular}
\end{subfigure}
\hfill
\begin{subfigure}[t]{0.245\textwidth}
\centering
\caption{Iter 1 ($G_1$)}
\label{fig:ncbi_gpt_iter1_1x4}
{\scriptsize \textbf{Gold} $\downarrow$ \hspace{0.4em} \textbf{Pred} $\rightarrow$ \par}
\vspace{0.15em}
\begin{tabular}{@{}c@{\hspace{0.45em}}c@{}}
\begin{tabular}{c|ccccc}
\toprule
 & C & D & M & S & O \\
\midrule
C & 0 & 0 & 0 & 0 & 0 \\
D & 0 & \cellcolor{red!22}1 & 0 & 0 & \cellcolor{red!22}1 \\
M & \cellcolor{red!22}1 & 0 & 0 & 0 & \cellcolor{red!34}2 \\
S & 0 & \cellcolor{red!70}6 & 0 & 0 & \cellcolor{red!34}2 \\
O & \cellcolor{red!22}1 & \cellcolor{red!70}6 & 0 & \cellcolor{red!22}1 & 0 \\
\bottomrule
\end{tabular}
&
{\scriptsize
\fbox{\begin{tabular}{@{}l@{}}
\textbf{TP}\\
C: 0\\
D: 12\\
M: 15\\
S: 33
\end{tabular}}}
\end{tabular}
\end{subfigure}
\hfill
\begin{subfigure}[t]{0.245\textwidth}
\centering
\caption{Iter 2 ($G_2$)}
\label{fig:ncbi_gpt_iter2_1x4}
{\scriptsize \textbf{Gold} $\downarrow$ \hspace{0.4em} \textbf{Pred} $\rightarrow$ \par}
\vspace{0.15em}
\begin{tabular}{@{}c@{\hspace{0.45em}}c@{}}
\begin{tabular}{c|ccccc}
\toprule
 & C & D & M & S & O \\
\midrule
C & 0 & 0 & 0 & 0 & 0 \\
D & 0 & \cellcolor{red!22}1 & 0 & 0 & \cellcolor{red!22}1 \\
M & \cellcolor{red!22}1 & 0 & 0 & 0 & \cellcolor{red!22}1 \\
S & 0 & \cellcolor{red!70}6 & \cellcolor{red!22}1 & 0 & \cellcolor{red!34}2 \\
O & 0 & \cellcolor{red!46}3 & 0 & \cellcolor{red!22}1 & 0 \\
\bottomrule
\end{tabular}
&
{\scriptsize
\fbox{\begin{tabular}{@{}l@{}}
\textbf{TP}\\
C: 0\\
D: 12\\
M: 15\\
S: 35
\end{tabular}}}
\end{tabular}
\end{subfigure}
\hfill
\begin{subfigure}[t]{0.245\textwidth}
\centering
\caption{Iter 3 ($G_3$)}
\label{fig:ncbi_gpt_iter3_1x4}
{\scriptsize \textbf{Gold} $\downarrow$ \hspace{0.4em} \textbf{Pred} $\rightarrow$ \par}
\vspace{0.15em}
\begin{tabular}{@{}c@{\hspace{0.45em}}c@{}}
\begin{tabular}{c|ccccc}
\toprule
 & C & D & M & S & O \\
\midrule
C & 0 & 0 & 0 & 0 & 0 \\
D & 0 & \cellcolor{red!22}1 & 0 & 0 & \cellcolor{red!22}1 \\
M & \cellcolor{red!22}1 & 0 & 0 & 0 & \cellcolor{red!22}1 \\
S & 0 & \cellcolor{red!58}5 & 0 & 0 & \cellcolor{red!34}2 \\
O & 0 & \cellcolor{red!46}3 & 0 & \cellcolor{red!22}1 & 0 \\
\bottomrule
\end{tabular}
&
{\scriptsize
\fbox{\begin{tabular}{@{}l@{}}
\textbf{TP}\\
C: 0\\
D: 12\\
M: 15\\
S: 33
\end{tabular}}}
\end{tabular}
\end{subfigure}

\caption{Discrepancy matrices over moderation iterations on NCBI Disease. Same-label cells indicate span-boundary mismatches, off-diagonal cells indicate remaining disagreement patterns, and TP counts are shown to the right of each matrix. C = CompositeMention, D = DiseaseClass, M = Modifier, S = SpecificDisease, O = No Entity.}
\label{fig:ncbi_gpt_iters_1x4}
\end{figure*}


\begin{figure}[t]
\centering
\small
\setlength{\fboxsep}{6pt}

\begin{subfigure}[t]{0.95\linewidth}
\centering
\fbox{%
\begin{minipage}{0.96\linewidth}
\begin{quote}
Homozygosity mapping in a family with\\
\underline{microcephaly}, \underline{mental retardation},\\
and \underline{short stature}
to a Cohen syndrome region on 8q21.3--8q.
\end{quote}
\end{minipage}}
\caption{Example of correct changes (resolved FNs).}
\label{fig:qual_fn}
\end{subfigure}

\vspace{0.5em}

\begin{subfigure}[t]{0.95\linewidth}
\centering
\fbox{%
\begin{minipage}{0.96\linewidth}
\begin{quote}
(WD) is an autosomal recessive disorder\\
characterized by \underline{copper accumulation} in \\
the liver, brain, kidneys, and corneas, and \ldots
\end{quote}
\end{minipage}}
\caption{Example of false changes (newly introduced FP).}
\label{fig:qual_fp}
\end{subfigure}

\caption{Example of correctly and falsely changed annotations after refinement.
(a) The three underlined entities correctly resolve original FNs, and 
(b) The underlined entity is a newly introduced FP.}
\label{fig:qual_examples}
\end{figure}


\section{Discussions}

\subsection{Performance Improvement Interpretation}
The simple prompt approach serves as a zero-shot annotation method that leverages the innate linguistic and world knowledge of LLMs.
While effective in domains where the model possesses sufficient background knowledge, performance on benchmark datasets often remains suboptimal.
This gap typically stems not from a deficiency in the model's underlying knowledge, but from a misalignment between the specific annotation conventions of the benchmark and the LLM's independent interpretation of the text.
Consequently, the F1-score improvements reported in Section~\ref{sec:results} suggest that
the use of guidelines successfully aligns the LLM's output with the rigorous requirements and boundary conventions encoded in the gold annotations.

\subsection{Limited Amount of Development Data}
The experiments are designed to simulate the moderation process typically conducted during the early stages of annotation projects,
where gold-standard supervision is highly limited.
This methodology represents a fundamental departure from traditional machine learning paradigms that rely on large-scale statistical learning.
While those empirical approaches require extensive datasets to capture representative statistics,
our work adopts a rationalistic approach that leverages the latent linguistic and world knowledge of LLMs.
Our primary hypothesis is that the advanced reasoning capabilities of modern LLMs can bridge
the gap between their internal knowledge and the specific conventions observed in concrete annotations.

This design choice also entails an important trade-off. The goal of the 10-document setting is not to approximate the statistically optimal amount of refinement data, but to approximate the practical conditions of early-stage guideline development, where only a small pilot set is typically available. 

At the same time, our experimental results also suggest that relying on only 10 documents as a basis for guideline refinement may be overly ambitious and limit the stability of the resulting improvements. First, such a small sample is highly susceptible to selection bias; there is a significant risk that substantial discrepancy patterns present in the wider dataset may not manifest within the sample.
Future research should examine the learning curve by incrementally increasing the volume of development data.
Second, certain components of the proposed method -- specifically the loop-termination criteria -- rely on statistical observations derived from this limited development set.
Because these metrics are calculated from a small sample size, they are inherently prone to instability.
Consequently, future work should focus on devising more robust or noise-tolerant termination conditions.

\subsection{Cost and Time Estimation}
Table~\ref{tab:moderation_process_cost_time_est} details the economic and temporal overhead of the moderation process. While GPT-5 demonstrates high performance, its operational cost is significantly higher, often by an order of magnitude compared to other models. Conversely, while DeepSeek offers the lowest financial cost per iteration, it suffers from significantly higher computational latency and lower overall performance.
Despite these variances, this suggests that LLM-based moderation has good potential to be a highly cost-effective alternative to human labor, although specialized human experts still retain an edge in absolute annotation quality.

\begin{table}[t]
\centering
\small
\setlength{\tabcolsep}{3.5pt}
\begin{tabular}{llcccccc}
\toprule
Dataset & Model & $i$ 
& $c_i$ (\$) & $t_i$ (min) 
& $\hat{C}_{proc}$ & $\hat{T}_{proc}$ \\
\midrule
\multirow{3}{*}{NCBI}
 & GPT-5    & 3 & 1.186 & 5.2  & 3.557 & 15.6 \\
 & Gemini   & 5 & 0.092 & 3.0  & 0.460 & 14.8 \\
 & DeepSeek & 2 & 0.054 & 15.8 & 0.109 & 31.6 \\
\midrule
\multirow{3}{*}{BC5CDR}
 & GPT-5     & 1 & 1.729 & 9.9  & 1.729 & 9.9 \\
 & Gemini    & 1 & 0.099 & 8.8 & 0.099 & 8.8 \\
 & DeepSeek  & 1 & 0.055 & 15.4 & 0.055 & 15.4 \\
\midrule
\multirow{3}{*}{BioRED}
 & GPT-5     & 2 & 1.991 & 14.0 & 3.982 & 28.0 \\
 & Gemini    & 1 & 0.251 & 5.9  & 0.251 & 5.9 \\
 & DeepSeek  & 1 & 0.048 & 29.8 & 0.048 & 29.8 \\
\bottomrule
\end{tabular}
\caption{
The cost of each iteration and the whole iterations of each experimental setup.
$c_i$ and $t_i$ denote the cost and time of the \emph{final} iteration $i$.
The end-to-end overhead is estimated as $\hat{C}_{proc}=i\cdot c_i$ and
$\hat{T}_{proc}=i\cdot t_i$.
}
\label{tab:moderation_process_cost_time_est}
\end{table}

\section{Conclusion}
\label{sec:conclusion}
This work investigated the systematic reuse and refinement of annotation guidelines within an LLM-based annotation paradigm. We empirically validated three core hypotheses regarding
(1) the efficacy of explicit guideline integration, 
(2) the advantage of reasoning-optimized architectures, and
(3) the viability of iterative moderation under minimum supervision.
While our results across multiple biomedical benchmarks provide support for all three hypotheses,
a granular analysis of discrepancy evolution against gold-standard annotations
suggests significant opportunities for further improvement.

\section*{Limitations}

Our approach assumes the availability of an initial human-written guideline document that is sufficiently detailed to support discrepancy-driven refinement.
In practice, many benchmarks and domains do not provide publicly available annotation guidelines, or only release high-level task descriptions, which limits the direct applicability of our framework. Addressing this limitation will require methods for \emph{guideline generation}, for example, by  LLM-consumable guidelines from annotation examples, label ontologies, or limited curator input.

In addition, while we evaluate refined guidelines extrinsically through strict span-and-type performance on held-out data, intrinsic assessment of guideline quality remains challenging. Guideline refinements may introduce unintended side effects or regressions, yet there is currently no standardized quality assurance procedure for evaluating guideline documents themselves. Developing principled \emph{guideline evaluation} and QA mechanisms, such as consistency checks or automated detection of overly broad rules, is an important direction for future work.

\section*{Ethical Considerations}

This work studies guideline-driven annotation and moderation using large language models
on existing, publicly available biomedical named entity recognition benchmarks (NCBI Disease, BC5CDR, and BioRED), which are released under the Creative Commons Attribution 4.0 (CC BY 4.0) license.
No new data collection or human subject involvement is required, and all experiments are
conducted on previously released datasets that are widely used in prior research.

The proposed framework does not introduce new predictive models or deploy systems in
real-world or clinical settings.
Instead, it focuses on analyzing systematic annotation discrepancies and refining
annotation guidelines in an offline, research-only context.
As such, the work does not pose additional ethical risks beyond those commonly associated
with benchmark-based NLP research.

\section*{Acknowledgments}

This work was supported by the Cross-ministerial Strategic Innovation Promotion Program (SIP) Second Phase, “Big-data and AI-enabled Cyberspace Technologies” by the New Energy and Industrial Technology Development Organization (NEDO).

In addition, we would like to express our gratitude to the Biomedical Linked Annotation Hackathon (BLAH) community for their feedback and advice.
\bibliography{custom}

\appendix
\section{Appendix}
\label{sec:appendix}

\paragraph{Inter-Annotator Agreement (IAA) Threshold}
To contextualize the termination criterion used in our moderation experiments, we report
previously published inter-annotator agreement (IAA) statistics for the benchmark datasets
considered in this work.
These values reflect the level of consistency typically achieved by trained human annotators
under the corresponding annotation guidelines.

For the BioCreative V Chemical--Disease Relation (BC5CDR) dataset, the reported IAA scores on the
test set are 87.49\% for disease entities and 96.05\% for chemical entities.
The BioRED dataset reports an entity-level IAA of 97.01\%, indicating high agreement among expert annotators.
The NCBI Disease corpus reports an inter-annotator agreement of approximately 87.5\% for disease mention annotation \citep{leaman2013dnorm}.

Considering these established human agreement levels, we set the moderation termination criterion to a strict-match F1 score of 0.9. This threshold is intentionally conservative relative to reported IAA values and is used solely as a stopping heuristic to prevent over-iteration during guideline refinement, rather than as a claim of exceeding or matching human annotation quality.

\begin{figure*}[!t]
\centering
\footnotesize
\setlength{\fboxsep}{5pt} 

\begin{subfigure}[t]{0.485\linewidth}
\centering
\fbox{%
\begin{minipage}{0.95\linewidth}
\texttt{%
\textbf{PROMPT TEMPLATE}\\
\\
\textbf{SYSTEM INSTRUCTION:}\\
You are an expert AI for text annotation. Your task is to annotate all entities from the provided text based on a strict schema and guidelines.\\
\\
\textbf{ENTITY SCHEMA:}\\
\{\{entitySchema\}\}\\
\\
\textbf{ANNOTATION GUIDELINES:}\\
\{\{guidelines\}\}\\
\\
\textbf{ANNOTATION RULES:}\\
- Output MUST be a valid JSON object with a single key "annotations".\\
- "annotations" MUST be an array of objects following this JSON schema:\\
\ \ \ \ \{\{jsonSchema\}\}\\
- If no entities are found, return \{"annotations": []\}.\\
- Spans must match the original text exactly. Do not alter spacing, casing, or punctuation.\\
\\
\textbf{TEXT TO ANNOTATE:}\\
---\\
\{\{inputText\}\}\\
---\\
\\
Provide your response as a single JSON object.%
}
\end{minipage}}
\caption{Prompt template.}
\label{fig:prompt_template}
\end{subfigure}
\hfill
\begin{subfigure}[t]{0.485\linewidth}
\centering
\fbox{%
\begin{minipage}{0.95\linewidth}
\texttt{%
\textbf{Biomedical Annotation Guidelines (Excerpt)}\\
\\
\textbf{What to Annotate}\\
\\
\textbf{1. Annotate all Specific Disease mentions}\\
A disease mention may refer to a \emph{Specific Disease} or a \emph{Disease Class}.\\
- Disease Class: A family of multiple specific diseases.\\
- Specific Disease: A single, well-defined disease entity.\\
\\
\emph{Example:}\\
Diastrophic dysplasia is an autosomal recessive disease.\\
Annotate ``Diastrophic dysplasia'' as Specific Disease and\\
``autosomal recessive disease'' as Disease Class.\\
\\
\textbf{2. Annotate contiguous text strings}\\
Composite mentions referring to multiple diseases are annotated as a single span.\\
\\
\textbf{3. Annotate disease mentions used as modifiers}\\
Disease names modifying other noun phrases are annotated as \emph{Modifier}.\\
\\
\textbf{4. Annotate duplicate mentions}\\
All disease mentions within a sentence are annotated, including duplicates.\\
\\
\textbf{5. Annotate the minimum necessary span}\\
Prefer the smallest span expressing the most specific disease form.\\
\\
\textbf{6. Annotate all synonymous mentions}\\
Long forms and abbreviations are annotated separately.\\
\\
\textbf{What NOT to Annotate}\\
- Organism names unless clearly referring to diseases.\\
- Gender terms unless defining a distinct disease subtype.\\
- Overlapping mentions.\\
- General terms (e.g., disease, syndrome), except cancer and tumor.\\
- Biological processes (e.g., tumorigenesis).\\
- Mentions interrupted by nested spans.\\
\\
\emph{Examples include composite mentions, disease classes, and exclusion cases.}%
}
\end{minipage}}
\caption{Human guideline excerpt.}
\label{fig:human_guidelines_example}
\end{subfigure}

\vspace{-0.3em}
\caption{Prompt template and Human guideline. (Left) Prompt template used for LLM-based annotation. (Right) Excerpt from the original human annotation guidelines - NCBI Disease Corpus.}
\label{fig:appendix_prompt_and_guidelines}
\end{figure*}

\begin{figure*}[t]
\centering
\small

\begin{tcolorbox}[
  width=0.97\textwidth,
  colback=green!5,
  colframe=green!40!black,
  boxrule=0.6pt,
  arc=2pt,
  left=6pt,right=6pt,top=6pt,bottom=6pt
]
\textbf{\#4. Annotate coordinated headless phenotype/pathology complements as Disease Class}

\vspace{0.4em}
When a coordinated sequence of \emph{headless} phenotype/pathology noun phrases (i.e., each item’s head is a pathological/phenotypic noun such as
\texttt{anemia}, \texttt{lymphadenopathy}, \texttt{splenomegaly}, \texttt{thrombocytopenia}, \texttt{rash})
functions as the complement of a clinical-description or diagnostic construction,
annotate \emph{each coordinated item} as \texttt{DiseaseClass} using the minimum contiguous span.

\vspace{0.6em}
\textbf{Clinical-description / diagnostic constructions include:}
\begin{itemize}[leftmargin=1.2em, itemsep=0.2em, topsep=0.2em]
  \item Phrases headed by or modifying a patient/disease/syndrome NP:
  \texttt{patients with ...}, \texttt{the disorder is characterized by ...},
  \texttt{a syndrome of ...}, \texttt{manifesting/presenting with ...},
  \texttt{features include ...}, \texttt{diagnosed with ...}.
  \item Constructions governed by a superordinate disease/syndrome head that introduces the list
  (e.g., \texttt{a syndrome characterized by ...}).
\end{itemize}

\vspace{0.4em}
\textbf{Annotate:}
\begin{itemize}[leftmargin=1.2em, itemsep=0.2em, topsep=0.2em]
  \item Each coordinated phenotype/pathology item as \texttt{DiseaseClass}.
  \item Use the smallest span that conveys the most specific pathological content
  (retain subtype-defining adjectives; avoid non-diagnostic intensifiers unless clinically defining).
\end{itemize}

\vspace{0.4em}
\textbf{Do not apply this rule when:}
\begin{itemize}[leftmargin=1.2em, itemsep=0.2em, topsep=0.2em]
  \item Items are bare general terms without specific pathological content (e.g., \texttt{complications}, \texttt{abnormalities}, \texttt{symptoms}).
  \item Items denote biological processes rather than disorders (e.g., \texttt{tumorigenesis}, \texttt{carcinogenesis}).
  \item The coordination depends on an overt disease head outside the items (e.g., \texttt{breast and ovarian cancer});
  in such cases follow \texttt{CompositeMention} rules instead.
\end{itemize}

\vspace{0.4em}
\textbf{Important.}
This rule adds \texttt{DiseaseClass} annotations for diagnostic/phenotypic lists and does not change how
\texttt{SpecificDisease} or disease \texttt{Modifier}s are annotated elsewhere (e.g., \texttt{ALPS} remains \texttt{SpecificDisease};
\texttt{ALPS phenotype} remains a \texttt{Modifier} and the generic head \texttt{phenotype} is not annotated).
\end{tcolorbox}

\caption{
Inserted guideline rule after moderation.
Full text of the newly added rule (\S4) that instructs annotating coordinated headless phenotype/pathology complements
as \texttt{DiseaseClass}, including scope conditions and exceptions.
}
\label{fig:appendix_guideline_rule4_latex}
\end{figure*}

\begin{figure*}[t]
\centering
\small
\setlength{\fboxsep}{6pt}

\begin{subfigure}[t]{0.485\textwidth}
\centering
\caption{\textbf{DiseaseClass FN (Iteration 0)}}
\label{fig:app_diseaseclass_before}
\fbox{%
\begin{minipage}{0.96\linewidth}
\begin{quote}
Homozygosity mapping in a family with\\
\textbf{microcephaly}, \textbf{mental retardation},\\
and \textbf{short stature}
to a Cohen syndrome region on 8q21.3--8q.
\end{quote}

\textit{Gold:} microcephaly, mental retardation, short stature (\textsc{DiseaseClass})\\
\textit{Prediction:} none (all missed) \textit{[FN]}
\end{minipage}%
}
\end{subfigure}
\hfill
\begin{subfigure}[t]{0.485\textwidth}
\centering
\caption{\textbf{DiseaseClass fixed (Iteration 1)}}
\label{fig:app_diseaseclass_after}
\fbox{%
\begin{minipage}{0.96\linewidth}
\begin{quote}
Homozygosity mapping in a family with\\
\underline{microcephaly}, \underline{mental retardation},\\
and \underline{short stature}
to a Cohen syndrome region on 8q21.3--8q.
\end{quote}

\textit{Gold:} microcephaly, mental retardation, short stature (\textsc{DiseaseClass})\\
\textit{Prediction:} all correctly recognized as \textsc{DiseaseClass}
\end{minipage}%
}
\end{subfigure}

\vspace{0.6em}

\begin{subfigure}[t]{0.485\textwidth}
\centering
\caption{\textbf{Correct non-annotation (Iteration 0)}}
\label{fig:app_wd_before}
\fbox{%
\begin{minipage}{0.96\linewidth}
\begin{quote}
(WD) is an autosomal recessive disorder\\
characterized by \textbf{copper accumulation} in the liver, brain, kidneys, and corneas, and \ldots
\end{quote}

\textit{Gold:} none\\
\textit{Prediction:} none (correct)
\end{minipage}%
}
\end{subfigure}
\hfill
\begin{subfigure}[t]{0.485\textwidth}
\centering
\caption{\textbf{FP introduced (Iteration 1)}}
\label{fig:app_wd_after}
\fbox{%
\begin{minipage}{0.96\linewidth}
\begin{quote}
(WD) is an autosomal recessive disorder\\
characterized by \underline{copper accumulation} in\\
the liver, brain, kidneys, and corneas, and \ldots
\end{quote}

\textit{Gold:} none\\
\textit{Prediction:} \textsc{DiseaseClass} (\underline{copper accumulation}) \textit{[FP]}
\end{minipage}%
}
\end{subfigure}

\caption{Detailed before/after examples across moderation iterations.
Top row: a coordinated \textsc{DiseaseClass} list is missed at iteration 0 but recovered after refinement (FN reduction).
Bottom row: ``copper accumulation'' is correctly unannotated at iteration 0 but becomes a false positive after refinement (precision--recall trade-off).}
\label{fig:detailed_examples}
\end{figure*}

\begin{table*}[t]
\centering
\small
\begin{tabular}{lccc}
\toprule
Model & S & G & M \\
\midrule
Llama3-8B   & 0.101 & 0.067 & 0.051 \\
Gemma3-12B  & 0.162 & 0.156 & 0.232 \\
Gemma3-27B  & 0.214 & 0.256 & 0.218 \\
\bottomrule
\end{tabular}
\caption{Preliminary pilot results for smaller local models. S = prompt-only, G = original guidelines, and M = guideline refinement. These models were excluded from the main experiments because they showed low baseline performance, unstable adherence to refined guidelines, and no consistent improvement.}
\label{tab:pilot_local_models}
\end{table*}

\begin{table*}[t]
\centering
\small
\begin{tabular}{llr}
\toprule
Dataset & Entity Type & Count \\
\midrule
NCBI Disease & CompositeMention & 37 \\
NCBI Disease & DiseaseClass & 127 \\
NCBI Disease & Modifier & 218 \\
NCBI Disease & SpecificDisease & 409 \\
\midrule
BC5CDR & Chemical & 1,195 \\
BC5CDR & Disease & 951 \\
\midrule
BioRED & CellLine & 22 \\
BioRED & ChemicalEntity & 822 \\
BioRED & DiseaseOrPhenotypicFeature & 982 \\
BioRED & GeneOrGeneProduct & 1,085 \\
BioRED & OrganismTaxon & 370 \\
BioRED & SequenceVariant & 250 \\
\bottomrule
\end{tabular}
\caption{Per-type entity counts in the 100-document evaluation sets.}
\label{tab:entity_counts}
\end{table*}

\begin{table*}[t]
\centering
\small
\begin{tabular}{llccccc}
\toprule
Dataset & Model & G & M & $\Delta$ & Bootstrap $p$ & AR $p$ \\
\midrule
NCBI
& GPT      & 0.7264 & 0.7588 & +0.0323 & 0.0762 & 0.0859 \\
& Gemini   & 0.6247 & 0.6548 & +0.0301 & 0.0408 & 0.0454 \\
& DeepSeek & 0.5521 & 0.5618 & +0.0097 & 0.3215 & 0.3159 \\
\midrule
BC5CDR
& GPT      & 0.8490 & 0.8610 & +0.0119 & 0.1198 & 0.1189 \\
& Gemini   & 0.7562 & 0.7722 & +0.0159 & 0.0350 & 0.0375 \\
& DeepSeek & 0.6400 & 0.6506 & +0.0106 & 0.2410 & 0.2462 \\
\midrule
BioRED
& GPT      & 0.7630 & 0.8165 & +0.0535 & 0.0002 & 0.00005 \\
& Gemini   & 0.6659 & 0.6891 & +0.0232 & 0.0410 & 0.0431 \\
& DeepSeek & 0.5333 & 0.5413 & +0.0079 & 0.2809 & 0.2896 \\
\bottomrule
\end{tabular}
\caption{One-sided significance tests for guideline refinement ($M > G$) under strict span-and-type matching. G = original guidelines, M = moderated guidelines, and $\Delta = M - G$. Bootstrap = paired document-level bootstrap resampling (5,000 iterations). AR = approximate randomization (20,000 permutations).}
\label{tab:significance_all}
\end{table*}

\begin{table*}[t]
\centering
\small
\begin{tabular}{llccc}
\toprule
Dataset & Model & Non-reasoning & Reasoning & $\Delta$ \\
\midrule
NCBI
& GPT      & 0.40 & 0.46 & +0.06 \\
& Gemini   & 0.38 & 0.40 & +0.02 \\
& DeepSeek & 0.08 & 0.31 & +0.23 \\
\midrule
BioRED
& GPT      & 0.67 & 0.74 & +0.07 \\
& Gemini   & 0.62 & 0.61 & -0.01 \\
& DeepSeek & 0.43 & 0.45 & +0.02 \\
\midrule
BC5CDR
& GPT      & 0.72 & 0.80 & +0.08 \\
& Gemini   & 0.69 & 0.68 & -0.01 \\
& DeepSeek & 0.55 & 0.65 & +0.10 \\
\bottomrule
\end{tabular}
\caption{Prompt-only comparisons between non-reasoning and reasoning variants. Across 9 comparisons, reasoning improves strict-match F1 in 7 cases, with an average gain of approximately +0.06 and only minor reversals for Gemini on BioRED and BC5CDR.}
\label{tab:prompt_reasoning}
\end{table*}

\end{document}